\documentclass[letterpaper, 10 pt, conference]{ieeeconf}

\IEEEoverridecommandlockouts                              

\usepackage[utf8]{inputenc}
\usepackage{multirow}
\usepackage{graphicx}
\usepackage{subcaption}
\usepackage{amsmath}

\usepackage{xcolor}

\definecolor{mygreen}{RGB}{0, 180, 0}

\newcommand{\fengyi}[1]{{\color{black} #1}}

\title{\LARGE \bf
	Object Classification Utilizing Neuromorphic Proprioceptive Signals in Active Exploration: Validated on a Soft Anthropomorphic Hand 
}

\author{Fengyi Wang$^{1}$, Xiangyu Fu$^{1}$ Nitish Thakor$^{2}$ and Gordon Cheng$^{1}$ 	
	\thanks{$^{1}$Fengyi Wang,  Xiangyu Fu, and Gordon Cheng are with the Institute for Cognitive Systems, Technical University of Munich, Arcisstrae 21, 80333 Munich, Germany {\tt\small \{ fengyi.wang, xiangyu.fu, gordon \} @tum.de}
	} %
	\thanks{$^{2}$Nitish Thakor is with the Department of Biomedical Engineering, Johns Hopkins University, Baltimore, MD, USA {\tt\small thakorjhu@jhu.edu}}
}
\begin{document}
	\maketitle

	\begin{abstract}
		Proprioception, a key sensory modality in haptic perception, plays a vital role in perceiving the 3D structure of objects by providing feedback on the position and movement of body parts. The restoration of proprioceptive sensation is crucial for enabling in-hand manipulation and natural control in the prosthetic hand. Despite its importance, proprioceptive sensation is relatively unexplored in an artificial system. In this work, we introduce a novel platform that integrates a soft anthropomorphic robot hand (QB SoftHand) with flexible proprioceptive sensors and a classifier that utilizes a hybrid spiking neural network with different types of spiking neurons to interpret neuromorphic proprioceptive signals encoded by a biological muscle spindle model. The encoding scheme and the classifier are implemented and tested on the datasets we collected in the active exploration of ten objects from the YCB benchmark. Our results indicate that the classifier achieves more accurate inferences than existing learning approaches, especially in the early stage of the exploration. This system holds the potential for development in the areas of haptic feedback and neural prosthetics.
	\end{abstract}

	\section{Introduction}
	
	In the human somatosensory system, both proprioception and tactile sensation play pivotal roles, enabling us to perceive tangible attributes of objects. Various types of primary afferent neurons transmit external stimuli into encoded spike trains, which are then relayed to higher levels of the haptic perception pathway. Tactile experiences often involve active exploration, such as grasping and lifting, to gather critical object information. In this process, proprioceptors provide information about the position and movement of one's own body parts, and motor signals convey information about intended movements \cite{gibsonObservationsActiveTouch1962}, \cite{ledermanExtractingObjectProperties1993}. It is believed that these proprioceptive and motor signals integrate with tactile signals at a higher perception level, culminating in the perception of an object's three-dimensional structure \cite{overvlietUseProprioceptionTactile2008}, \cite{robles-de-la-torreForceCanOvercome2001}. Compared to the sense of touch, proprioception, despite its significance, has not been extensively investigated in a neuromorphic system for object discrimination tasks.
	
	Humans can distinguish objects within a very short time upon contact. Fast feedback or decision-making is also crucial for some applications, such as neural prostheses that use spiking signals for nerve stimulation and provide sensory feedback to human users \cite{oddoIntraneuralStimulationElicits2016},  \cite{osbornProsthesisNeuromorphicMultilayered2018}. A classifier that can make accurate decisions based on information obtained in a short period of time is beneficial. We draw inspiration from the biological mechanism whereby the signals are encoded and transmitted as spikes in the nervous system and propose an object classifier that utilizes a hybrid spiking neural network (SNN) to process the proprioceptive signals. The sensory readings are encoded into spike trains with the mathematical model of the muscle spindle and transmit information exclusively via spikes. In this approach, the energy consumed by data transfer in artificial neural networks (ANNs) can be significantly reduced, enabling powerful computations with a lower power budget, especially when implemented on neuromorphic chips \cite{orchardEfficientNeuromorphicSignal2021}. \fengyi{In this work, we implemented an SNN-based classifier to distinguish 10 different objects from the YCB benchmark.}
	
	Anthropomorphic robotic devices offer an excellent platform for these studies. They are designed to resemble and mimic human body parts, often both in form and function. In this work, we employed the QB SoftHand that mimics the mechanism of the human hand. An essential aspect of the hand's design is its reliance on adaptive synergies \cite{catalano2014adaptive}, a concept in underactuated hand design that leverages principles of human hand motion. Coupled with its distinctive flexible joint design, the QB SoftHand can be controlled by a single motor and tendon system yet achieves a wide variety of grasps due to its intrinsic adaptability to object shapes. 
	The hand only provides the position and current of its motor as feedback, but the underactuated design and the adaptive joints make it infeasible to determine the configuration of fingers merely by the position of its actuator. To address these limitations, we integrated multiple stretch sensors with the hand to provide proprioceptive feedback. The adaptability also allows it to sense the shape and the weight of an object through the passive movement of the fingers in active explorations. This approach facilitates the analysis of the biological encoding scheme and the perception pathway by testing their functional performance in robotic tasks.
	
	\begin{figure*}
		\centering
		\includegraphics[width=0.7\linewidth]{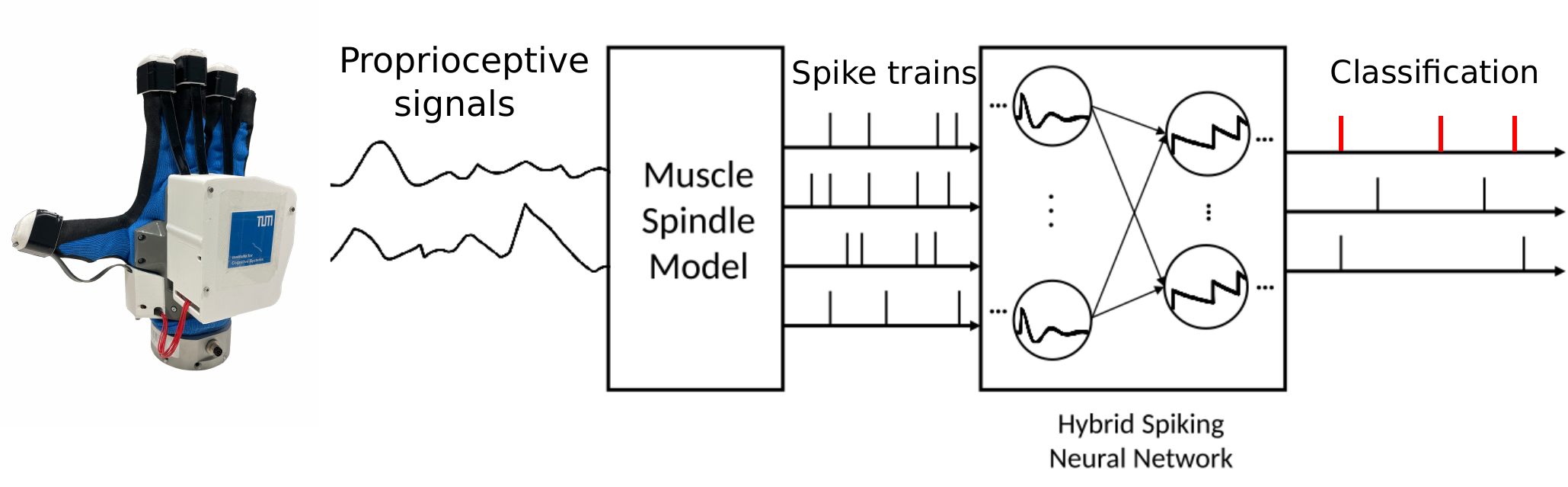}
		\caption{The schematic diagram of the system. The proprioceptive sensor readings are encoded into spike trains with a biological muscle spindle model and fed to a hybrid spiking neural network for classification. The hybrid SNN includes a resonator layer and multiple integrator layers.}
		\label{fig:system}
	\end{figure*}
	
	In summary, the main contributions of this paper are:	
	
	\begin{itemize}
		\item Integration of stretch sensors with the QB Softhand to create a hardware platform that facilitates research into proprioceptive perception;
		\item Development of a neural encoding scheme that integrates signals from proprioceptive sensors based on the mathematical model of proprioceptors;
	\item Implementation of a \fengyi{hybrid SNN-based classifier in an object classification task}. This classifier combines the advantages of integrator and resonator neurons, proving to be more accurate than other popular machine learning approaches in classifying ten objects through active exploration.

	\end{itemize}

	\section{Background and Related Work}

	Haptic perception is an active research field, encompassing areas such as object classification \cite{hosodaRobustHapticRecognition2010}, \cite{kerzelNeuroRoboticHapticObject2019}, object manipulation \cite{jamesSlipDetectionGrasp2021}, and texture discrimination \cite{sankarTextureDiscriminationSoft2021}. There was also research dedicated to providing sensory feedback to amputees through prosthetics \cite{osbornProsthesisNeuromorphicMultilayered2018}.
	
	Tactile signals provide us with rich information about objects, such as their edges and surface textures \cite{thakur2006receptive}, \cite{saalTouchTeamEffort2014}. However, humans are still able to discriminate among many different objects based on their 3D structure, even when tactile sensations are largely diminished, such as when wearing gloves. Compared to tactile signals, proprioceptive feedback has received less attention despite its significance \cite{overvlietUseProprioceptionTactile2008}.
	
	Anthropomorphic robotic hands offer a platform for studying the use of proprioceptive feedback in humans in an artificial system. Kaboli and Cheng \cite{RobustMohsen} achieved high accuracy in discriminating textures and objects by utilizing a robust tactile descriptor and intricate manipulation with the Shadow Dexterous Hand, which is integrated with fingertip tactile sensors. 
	
	Although these devices are trying to mimic the form and function of a human hand, the driving mechanisms and sensory modalities of many anthropomorphic hands and humans are still quite different. For example, the Shadow Dexterous Hand is driven by independent actuators within joints and provide precise joint positions as proprioceptive feedback for determining the configurations of each finger segment. In contrast, soft biomimetic robotic hands, which have similar driving mechanisms as the human hands, often lack proprioceptive feedback and rely solely on tactile sensing for perception tasks such as object classification  \cite{ward-cherrierMiniaturisedNeuromorphicTactile2020} and texture discrimination \cite{sankarTextureDiscriminationSoft2021}. 
	
	Human fingers are driven by muscles and tendons and typically exhibit synergistic actions, and the most important kinesthetic feedback regarding the length and velocity of the muscle is provided by the spindle embedded in it \cite{hulligerMammalianMuscleSpindle1984}, \cite{matthewsEvolvingViewsInternal1981}. There have been many attempts to develop models of the muscle spindle. Among these, Vannuci et al. \cite{vannucciProprioceptiveFeedbackNeuromorphic2017} modified the mathematical model by Mileusnic et al. \cite{mileusnicMathematicalModelsProprioceptors2006} for real-time neuromorphic applications. This model receives the fascicle length and the $\gamma$-motor activation level to generate the spiking activities of the primary and secondary afferents of the muscle spindle, incorporating all three types of intrafusal fibers within the spindle.
	
	This mathematical model was employed by Rostamian et al. \cite{rostamianTextureRecognitionBased2022} to encode the motion of the robotic arm into spike trains, which were then fed into a classifier for the texture recognition task. However, the proprioceptive signal was only utilized to achieve scanning speed-invariant results. The finger movement elicited by external stimuli was not mediated by proprioceptive signals.
	
	Most previous work uses conventional machine learning models such as k-nearest neighbor (kNN) or supporting vector machine (SVM) to classify the encoded spike trains. In these approaches, features such as average interspike intervals (ISI) \cite{sankarTextureDiscriminationSoft2021} are extracted from the spike train. These features are usually hand-crafted and highly task-specific. Furthermore, these approaches need to wait for time windows to extract enough features, causing significant delays in decision-making.
	
	The spiking neural network is designed for processing spike trains and has demonstrated its power efficiency on neuromorphic hardware. Taunyazov et al. \cite{taunyazovFastTextureClassification2020} employed SNN to classify textures in two tactile datasets obtained through sliding motion. They successfully demonstrated that SNNs have an advantage over SVMs in terms of accuracy, particularly in the early stages of the stimuli. However, when the complete data sequence is available, the accuracy of SNN was surpassed by that of artificial neural networks (ANNs). This result, as our testing on the same dataset indicates, is attributed to the simple threshold-crossing encoding method used.
	
	Most SNNs employ integrator neurons that perform temporal integration of the incoming pulse trains, such as the widely used leaky-integrate-and-fire neuron. In contrast, resonant behavior has been observed in some neurons in the perception pathway  \cite{gutfreundSubthresholdOscillationsResonant1995}. Such neurons prefer inputs having a certain resonant frequency that is equal to the frequency of the subthreshold oscillation \cite{izhikevichResonateandfireNeurons2001}.
	
	In this study, our emphasis is on object classification tasks, specifically with proprioceptive signals. We implemented the neuromorphic encoding scheme for proprioceptive measurements and collected a dataset with human-like active exploration. A hybrid SNN is then constructed and demonstrates the ability for more accurate classification.
	
	\section{Methods}
	
	\begin{figure}
		\centering
		\includegraphics[width=0.6\linewidth]{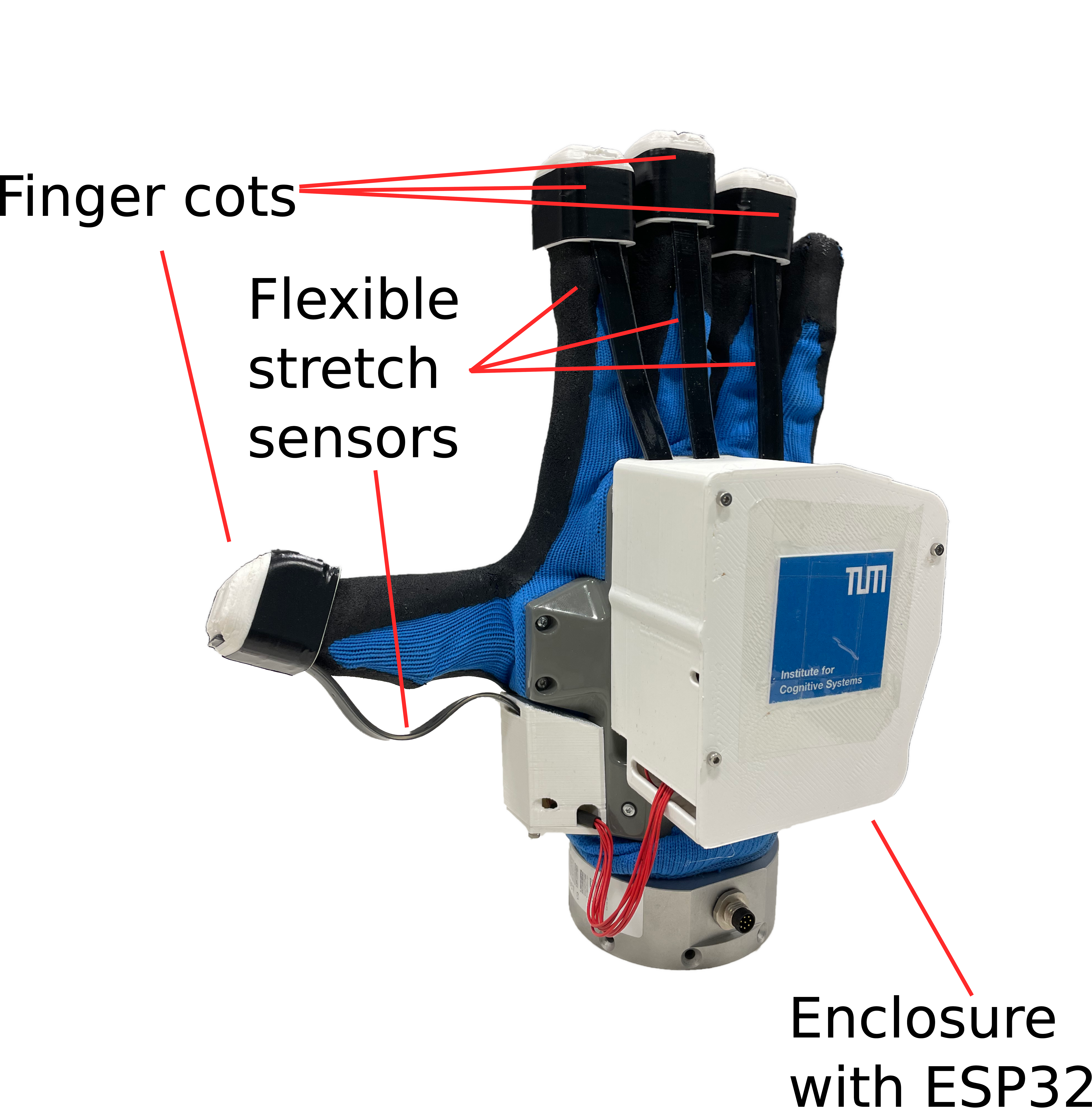}
		\caption{The QB Softhand integrated with four flexible stretch sensors on the dorsal side of the thumb, index finger, middle finger, and ring finger, respectively.}
		\label{fig:handfrontandback}
	\end{figure}
	
	\subsection{Proprioceptive sensor} \label{mtd:proprio_sensor}
	The QB SoftHand is only capable of sensing the motor position and the magnitude of the motor current, lacking feedback on the actual pose of each individual finger. Therefore, we installed stretch sensors from Bend Labs on the dorsal side of the thumb, index finger, middle finger, and ring finger. One end of the sensor is fixed inside a 3D-printed enclosure mounted on the dorsal side of the QB Softhand, while the other end is attached to the corresponding finger cot as shown in Fig. \ref{fig:handfrontandback}.

	Since human proprioception primarily relies on muscle spindle signals that encode muscle length and motor activation level rather than joint angles, in this work, we only utilized the stretch measurement, even though the sensor also provides path-independent angular displacement measurement.

	\subsection{Neural Encoding of Proprioceptive signals}
	
	In this work, we use an ESP32 to read the sensor measurements at 350Hz and encode them in real time into spike trains. We adapted Vannuci's neuromorphic muscle spindle model \cite{vannucciProprioceptiveFeedbackNeuromorphic2017} to encode signals from the proprioceptive sensors that we mentioned in the previous section. The model encodes relative fascicle length $L$ and its rate of change $\dot{L}$ into spiking activities of primary (Ia) and secondary afferent (II). This process also considers the relative activation levels of static ($\gamma_s$) and dynamic ($\gamma_d$) gamma motor axons. We mapped the sensor's minimum and maximum reading in the experiment to $[0.95L_o, 1.08L_o]$ to ensure consistency with the model.
	
	The primary afferents transmit information about both $L$ and $\dot{L}$, whereas the secondary afferent predominantly conveys information regarding $L$.  The spiking activity of primary and secondary afferents are related to the tension of all three types of intrafusal muscle fibers: long nuclear Bag$_1$ and Bag$_2$ fibers and shorter chain fibers. The fiber tension $T(t)$ can be computed by performing a discrete fixed-step integration. 
	
\begin{figure}
		\centering
		\includegraphics[width=1\linewidth]{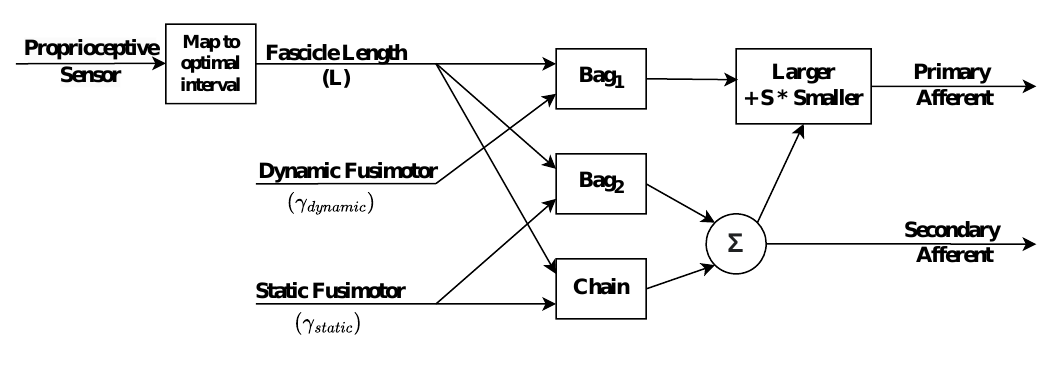}
		\caption{The schematic diagram of the muscle spindle model, which consists of 3 intrafusal fiber models.}
		\label{fig:blockdiagramm}
	\end{figure}

	The smaller response between Bag1 and the sum of Bag2 and Chain is then multiplied by a factor of $S = 0.156$. The total response of all fibers represents the primary afferent response, as shown in Fig. \ref{fig:blockdiagramm}.
	
	The spike rate of $\gamma_s$ and $\gamma_d$ are set to 70 pulses per second to maintain the overall sensitivity of both primary and secondary afferents, providing sensory feedback even when the muscle is contracted. 	For more details, please refer to \cite{vannucciProprioceptiveFeedbackNeuromorphic2017} and \cite{mileusnicMathematicalModelsProprioceptors2006}.
	
	\begin{figure}[]
		\centering
		\includegraphics[width=1\linewidth]{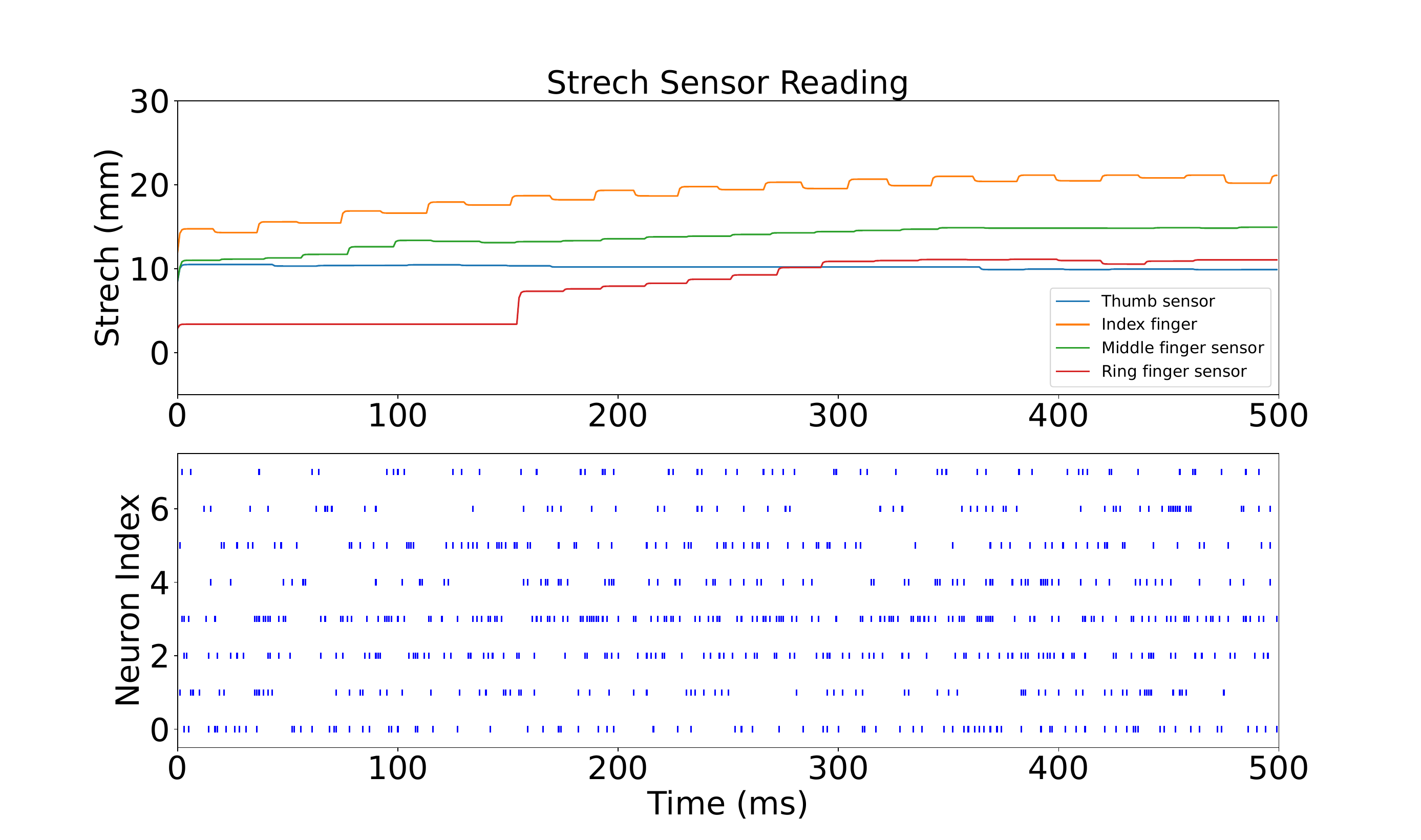}
		\caption{Encoding of proprioceptive signal with muscle spindle model while the hand is closing.}
		\label{fig:signal}
	\end{figure}
	
	A homogeneous Poisson process \cite{heeger2000poisson} is used to generate a spike train from the resulting instantaneous spike rate $r$:
	\begin{equation}
		P\{\text { spike during } \Delta t\}=r \cdot \Delta t
	\end{equation}
	where $\Delta t = 1 \mathrm{~ms}$ is the duration of a simulation time step. This approach was suitable because we had fixed time bins $\Delta t\ $, which is much smaller than the update interval of the sensors. An illustration of the encoded spike trains is shown in Fig. \ref{fig:signal}. 
	
		\begin{figure*}
		\centering
		\includegraphics[width=0.8\linewidth]{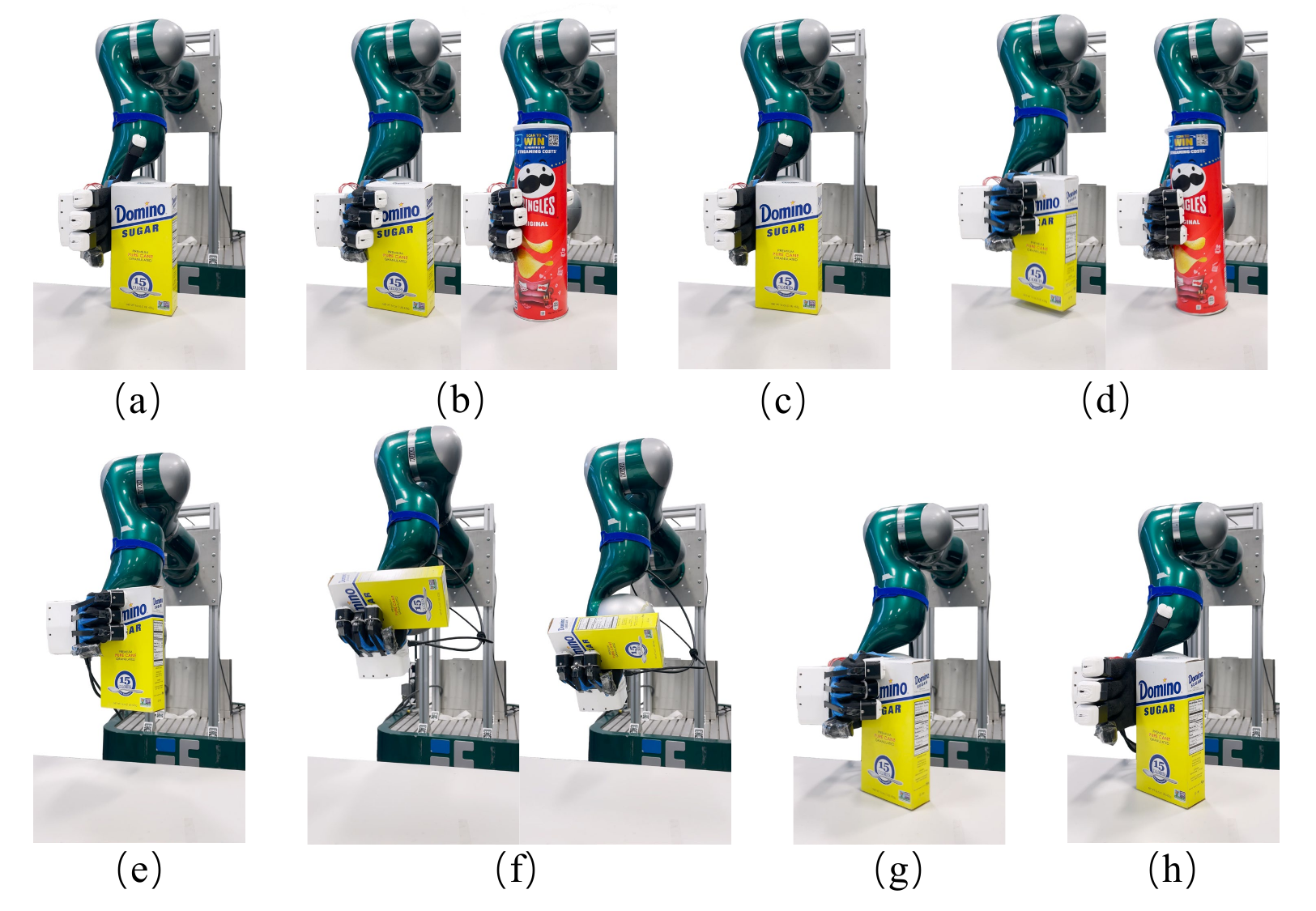}
		\caption{Illustration of a whole active exploration process in data collection. (a) shows the initial setup. (b) and (d) compares smaller and larger objects in light and strong grasp. (c) shows the release after the light grasp. (e) and (f) illustrate the lift-and-weigh stage in the exploration. In (g) and (h), the object is put down for the next trail.}
		\label{fig:grasp_process_bendlabs_only}
	\end{figure*}
	\subsection{Active exploration with the QB Softhand}

	The QB SoftHand integrated with the sensors is mounted on the 7-DOF KUKA LBR iiwa robotic arm as shown in Fig. \ref{fig:grasp_process_bendlabs_only}. The entire setup is managed through the Robot Operating System (ROS) to allow for more precise control and autonomous movements.
	
	Humans employ a variety of actions to actively explore their environment, such as trying to extract more features of objects through in-hand manipulation. Due to the limitations of the QB soft hand, which is driven by a single actuator through synergy, it's unable to complete complex in-hand manipulations similar to those of the Shadow Dexterous Hand \cite{RobustMohsen}. However, we can still accomplish a set of human-like active exploration movements through the cooperation of the hand and the robotic arm.

	We applied active exploration that can be divided into the light-heavy grasp stage and the lift-and-weigh stage. The light-heavy grasp phase focuses on perceiving objects' different shapes and stiffness, while the grasp-weigh action concentrates on the weights of objects by sensing their passive movements of the fingers.

	The active exploration process is illustrated in Fig \ref{fig:grasp_process_bendlabs_only}.  In each trail, The object is  positioned in front of the hand a light grasp and release are applied, followed by a stronger grasp. In the case of objects that are smaller or possess more complex shapes, like a sugar box and pliers, the robotic hand generated distinct configurations for light and heavy grasps, respectively. Then, the robotic arm lifts the object. The hand then rotates to position the palm upward, moves up and down twice to weigh the object through passive finger movements, and finally restores the object to its original location.

	\subsection{Data Collection}

\begin{figure}
		\centering
		\includegraphics[width=1\linewidth]{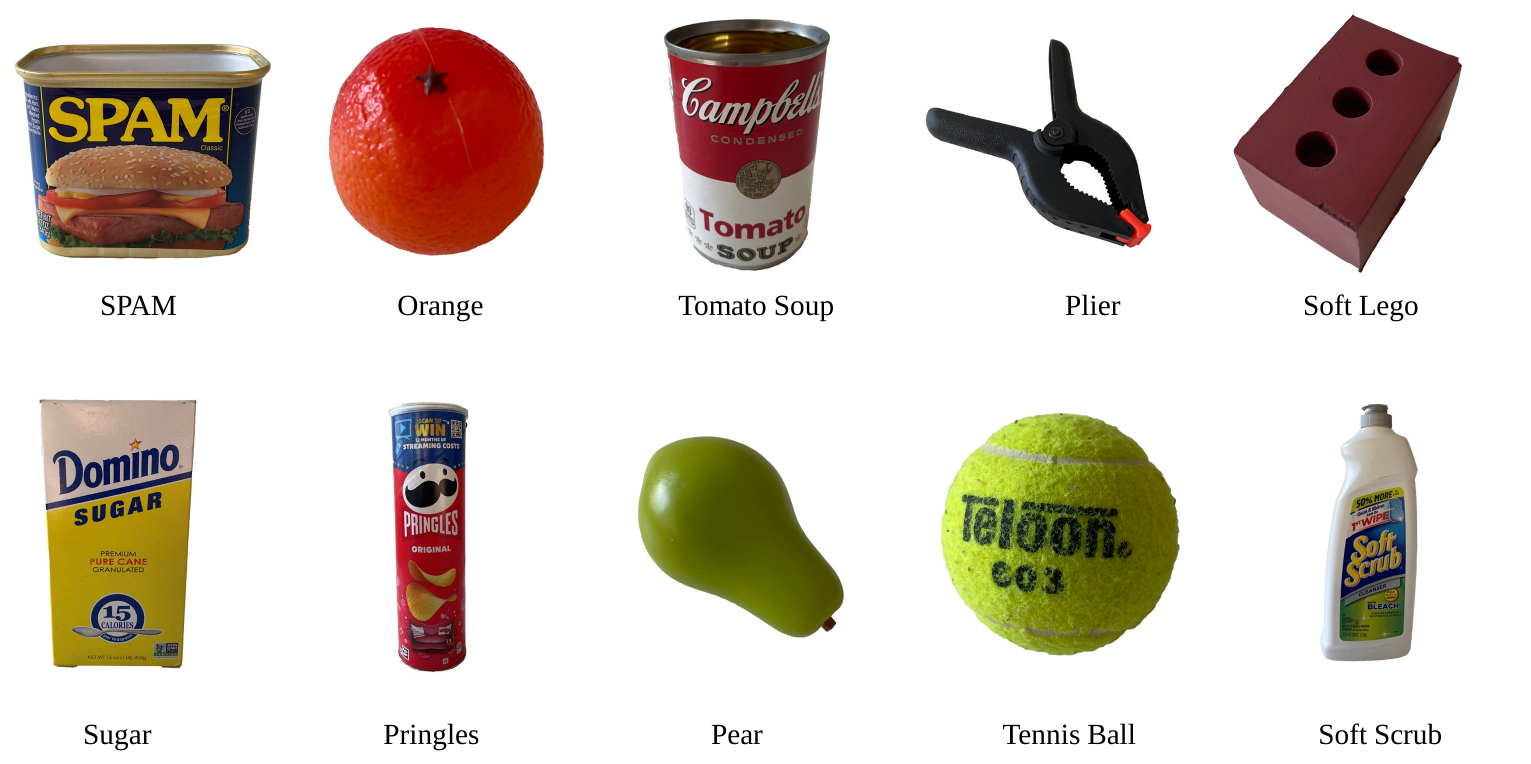}
		\caption{The objects used for dataset collection from the YCB benchmark. This dataset includes objects with similar shapes, such as oranges and tennis balls, tomato soup, and Pringles. It also includes soft objects like the soft Lego.}
		\label{fig:datasetwithnames}
	\end{figure}

	In each trial, we recorded the signals from the start of the grasp until 100 ms after the completion of the lift-and-weigh stage. With this setup, ten different objects from the YCB benchmark are collected with 100 samples each. \fengyi{In this process, objects are presented in similar positions.} The objects we used are shown in Fig.\ref{fig:datasetwithnames}. Since we use two afferents from each spindle model to encode the proprioceptive signal from a sensor, the encoded spike trains have eight channels. The total sequence length of each sample is 6000 (6s sampled at 1000 Hz).
	
	\subsection{Neuromorphic Classifier and Training}

	Training with event-based data presents a significant challenge because the spike generation function's derivative is undefined. In this study, we utilize the Spike Layer Error Reassignment in Time (SLAYER) algorithm \cite{shresthaSLAYERSpikeLayer2018} to train the SNN. SLAYER adopts a temporal credit assignment approach, enabling the back-propagation of errors to precede layers for updates. The SNN training is conducted on conventional GPU. After training, the SNN can be executed at high speed on a GPU to meet the requirements of online inference. Additionally, it can also be deployed on neuromorphic hardware for fast and power-efficient inference.
	
	We use a 4-layer fully connected architecture for our SNN model. The first layer of this neural network consists of Izhikevich resonate-and-fire (IZ-RF) neurons \cite{izhikevichResonateandfireNeurons2001}. Meanwhile, the other layers employed current-based leaky integration and fire (CUBA) neurons. The sizes of both hidden layers are set to 400. The dynamic of the CUBA neuron follows Equation (\ref{eq:cuba}).
	
	\begin{equation}
		\label{eq:cuba}
		\begin{aligned}
			x(t)&=\sum w s(t-1)\\
			i[t] &= (1 - \alpha_i)\,i[t-1] + x[t] \ \\
			v[t] &= (1 - \alpha_v)\,v[t-1] + i[t]  \\
		\end{aligned}
	\end{equation}	
	where $w$ is the input synaptic weights of the neuron, $s$ is the input spike. Neuron's input current $i[t]$ is modulated by the weighted sum of input spike $s_i[t]$ and a decay factor $\alpha_i$. The membrane potential $v[t]$ is then updated by $i[t]$ and its decay factor $\alpha_v$. When $v[t]$ exceeds the threshold $\theta$, the neuron emits a spike and resets its membrane potential $v[t]$ to zero.
	
	In contrast, the resonate neuron exhibits damped oscillation activity of the membrane potential. This activity is described with a complex-valued variable $z=x+\mathrm{i} y \in \mathbf{C}$ and a damping factor $\alpha$. $\phi$ defines the frequency of the oscillations. The input spikes are weighted by a complex-valued synaptic weight $w_r\in \mathbf{C}$. The sub-threshold behavior of an IZ-RF neuron follows Equation (\ref{eq:res}).
	
	\begin{equation}
		\label{eq:res}
		\begin{aligned}
			\mathrm{Re}(z[t]) &= (1-\alpha)(\cos\phi\ \mathrm{Re}(z[t-1])
			\\ &- \sin\phi\ \mathrm{Im}(z[t-1])) + \mathrm{Re}(w_r)s_i[t]\\
			\mathrm{Im}(z[t]) &= (1-\alpha)(\sin\phi\ \mathrm{Re}(z[t-1]) \\ &
			+ \cos\phi\ \mathrm{Im}(z[t-1])) + \mathrm{Im}(w_r)s_i[t]
		\end{aligned}
	\end{equation}
	
	When the imaginary state is above the threshold, the neuron fires and the real state is reset to zero.

	
	The input to the neural network is the encoded spike train, and the output is also the spike train with the size of a number of classes. The classifier makes decisions by looking at the neuron that emits the highest number of spikes in the output layer. The parameters of the neurons are listed in Table \ref{tab:neuron}.
	
	The desired output counts for neurons representing the correct class (positive spike counts) and the other classes (negative spike counts) are required by the SLAYER 2.0 algorithm in the LAVA software framework \cite{LavaSoftwareFramework}. We set the ratio of positive and negative spike counts to 0.5 and 0.1 empirically.
	
	\begin{table}[]
		\centering
		\caption{Neuron parameters}
		\label{tab:neuron}
		\resizebox{0.9\columnwidth}{!}{%
			\begin{tabular}{cccc}
				Neuron type & threshold & current\_decay & voltage\_decay \\ \cline{2-4} 
				IZ-RF       & 1.25      & 0.25           & 0.03           \\
				& threshold & period         & decay          \\ \cline{2-4} 
				CUBA        & 1         & 10             & 0.02          
			\end{tabular}%
		}
	\end{table}

	\subsection{Comparison Models}
	\textbf{ANN models:} We implemented a neural network includes Long-Short Term Memory (LSTM) units \cite{hochreiterLongShortTermMemory1997}, known for their proficiency in learning and continuously predicting temporal sequences. Each LSTM unit is characterized by a cell state and a hidden state, which are updated based on the temporal patterns of the input data. The models process four-channel proprioceptive signals as inputs. \fengyi{The ANN also has the fully connected three-layered structure as the SNN.}

	\textbf{kNN models}: kNN models are employed to assess how effectively encoded spike trains represent the proprioceptive signals. These models, utilizing a linear kernel, operate on both the raw sensory data along the time dimension and the encoded spike trains. In the second approach, we explored the utility of rate coding, where inputs are depicted as the spike count within a 100ms sliding time window, to distinguish between different classes.

	\section{Results}
	
	\subsection{Classification results}

	We use 80\% of proprioceptive data set for training. Each neural network is trained for 1000 epochs 5 times. The mean accuracy and standard deviations of different classifiers are summarized in Table \ref{tab:acc}.
	
	While the SNN models outperformed the other models, the hybrid SNN effectively prevails in terms of accuracy compared to an SNN of the same size using only CUBA neurons, although the training is slightly slower.
	
	The kNN models on the raw sensory data are far inferior to the proposed classifier. In contrast, the kNN models operated on the spiking data had much better performance than the kNN models on the raw sensory data.  This indicates that the neuromorphic encoding can effectively extract features in the time domain from the raw data. 
	
	\begin{table}[b]
		\centering
		\caption{Object Classification Accuracy Scores}
		\label{tab:acc}
		\resizebox{0.7\columnwidth}{!}{%
			\begin{tabular}{c|c}
				Models      & Accuracy \\ \hline
				Proposed SNN (hybrid)         & \textbf{79.3\%} (0.002)   \\
				SNN (CUBA)         & {70.2\%} (0.006) \\
				LSTM (Raw)        &    58.5\% (0.056)      \\
				kNN (Raw)   & 41.75\% (0.047)           \\
				kNN (Spike)   &    67.65\% (0.028)     \\
			\end{tabular}%
		}
	\end{table}
	\subsection{Early stage classification}

	\begin{figure}
		\centering
		\includegraphics[width=0.8\linewidth]{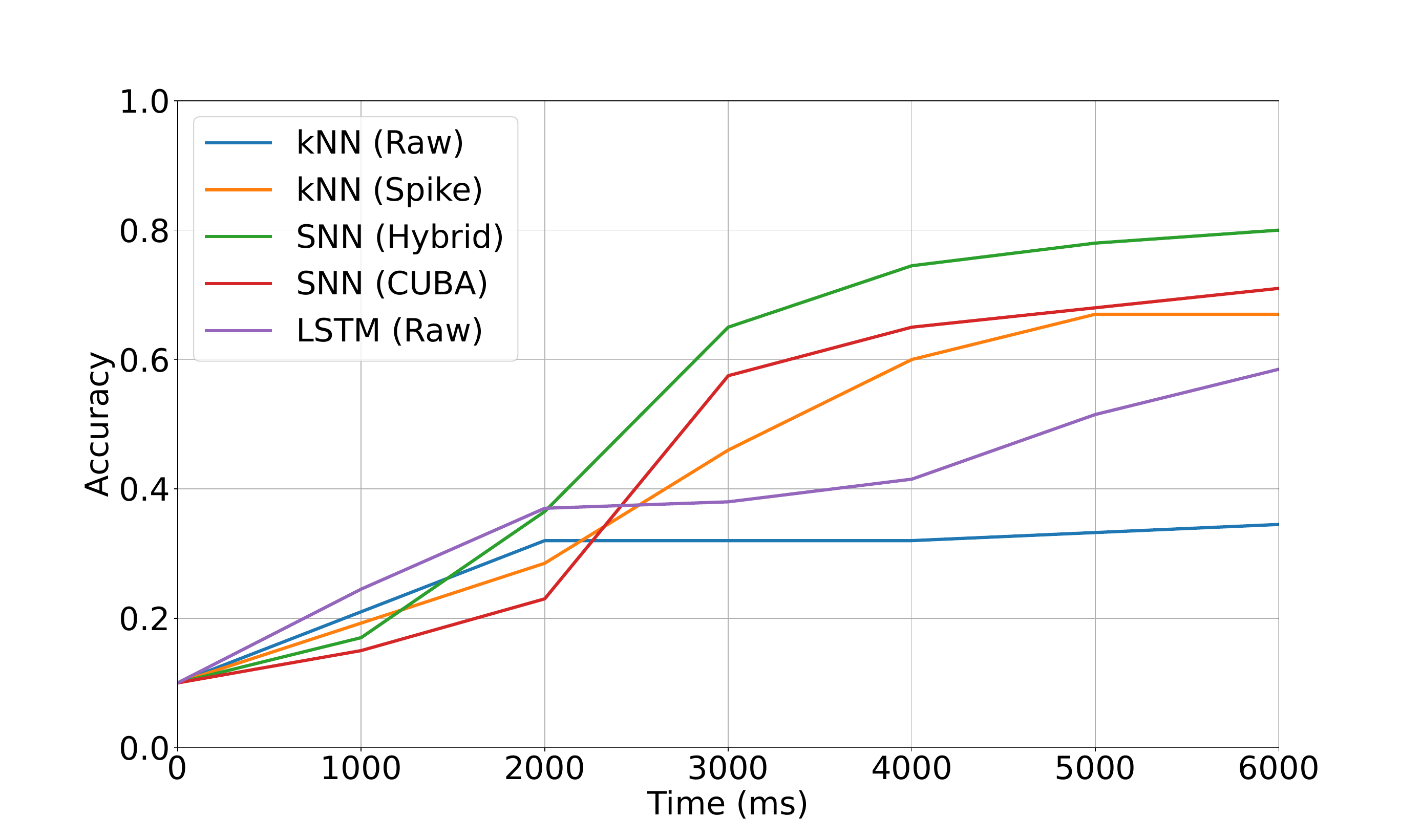}
		\caption{Accuracy score over time using different models. The proposed SNN (green) achieves an accuracy during the early stages of exploration that other models only reach when utilizing the full time series.}
		\label{fig:fast_class}
	\end{figure}
	
	One main advantage of SNN is the capability to make more accurate inferences at the early stage of the exploration, which is crucial for robotic applications. As shown in Fig. \ref{fig:fast_class}, the proposed classifier achieved 65\% accuracy, which is comparable to the highest accuracy of the comparison models. The proposed classifier can update at a frequency of 1000 Hz on an Nvidia RTX 4080 GPU. This makes it suitable for real-time applications.

Both kNN models need to incorporate data within a specified window as features, and the size of the window needs to be manually defined and is highly task-specific. The larger the time window, the greater the delay in making decisions and taking subsequent actions. SNNs are not posed to these issues because they can make inferences when the spike signal arrives.

	\section{Conclusion and future work}
%
	
	In this work, we integrated proprioceptive sensors with the QB Softhand and proposed an object classifier using a hybrid SNN. We used a mathematical model of muscle spindles to generate spike trains for the dataset we collected on ten objects from the YCB dataset. This encoding can be used to train machine learning models, such as SNN or k-Nearest Neighbors (kNN), to improve classification accuracy.
	
	\fengyi{This work does not yet include research on the pose invariance of the method. Future work could address this limitation by exploring active exploration strategies,} which involve adjusting the hand's movements based on sensory feedback. Additionally, investigating the role of $\gamma$-motor activation in perception and integrating tactile and proprioceptive signals could further enhance object classification accuracy and system performance.
	
	\section{Acknowledgement}
	This work was supported by the German Federal Ministry of Education and Research (BMBF) under Grant 01GQ2108.
	

	\bibliographystyle{IEEEtran}
	\bibliography{ref}

\end{document}